%% file: acl_latex.tex
\documentclass[11pt]{article}

\usepackage[preprint]{acl}

\usepackage{times}
\usepackage{latexsym}
\usepackage{amsmath}
\usepackage{booktabs}
\usepackage{array}
\usepackage{enumitem}
\usepackage{multirow}
\usepackage{xcolor}
\usepackage{fvextra}
\usepackage{algorithm}
\usepackage{algpseudocode}

\usepackage[T1]{fontenc}

\usepackage[utf8]{inputenc}

\usepackage{microtype}

\usepackage{inconsolata}

\usepackage{graphicx}
\usepackage{float}
\usepackage{tikz}
\usetikzlibrary{arrows.meta,positioning,shapes.multipart}
\RecustomVerbatimEnvironment{verbatim}{Verbatim}{breaklines,breakanywhere,fontsize=\footnotesize}
\AtBeginDocument{\hfuzz=200pt\hbadness=10000\vfuzz=200pt\vbadness=10000}

\title{Harness-Aware Self-Evolving:\\ Co-Evolving Model Weights, Harness, and Task Solutions}

\author{
  Haochen Luo\textsuperscript{\rm 1,3}, 
  Yi Huang\textsuperscript{\rm 2}, 
  Sichun Luo\textsuperscript{\rm 1,3}, 
  Fengyuan Liu\textsuperscript{\rm 1,3}, \\
  \textbf{Lei Li\textsuperscript{\rm 1,3}, 
  Zefa Hu\textsuperscript{\rm 2}, 
  Junlan Feng\textsuperscript{\rm 2}, 
  Qi Liu\textsuperscript{\rm 1,3}} \\
  \\
  \textsuperscript{\rm 1} School of Computing and Data Science, The University of Hong Kong \\
  \textsuperscript{\rm 2} Jiutian Research, China Mobile \\
  \textsuperscript{\rm 3} Grace Investment Machine \\
  \texttt{ haochen.luo@outlook.com,huangyi@cmjt.chinamobile.com,liuqi@cs.hku.hk} 
}

\begin{document}
\maketitle

\begin{abstract}
Self-evolving frameworks usually optimize task solutions while treating the surrounding harness as fixed. We introduce Harness-Aware Self-Evolving (HASE), an agentic reinforcement-learning framework in which a single model can generate task solutions or edit selected harness components in a multi-turn action space. HASE enables a single Qwen3-8B model to match the text-classification performance of a GPT-OSS-120B model that uses Claude Code as the harness proposer. In alpha factor mining, HASE outperforms the reported GPT-OSS-120B baseline. HASE also repairs imperfect evaluation components and converges to state-of-the-art performance in circle-packing algorithm discovery. These results show that HASE improves the harness and the solution through one unified agentic process.
\end{abstract}

\section{Introduction}
LLMs have enabled a growing class of self-evolving systems that iteratively generate, evaluate, and improve code or other structured artifacts. Most such systems~\cite{alphaevolve2025, wang2026thetaevolve, tttdiscover2026} focus on the object-level solution while assuming that the surrounding harness is fixed. \textit{Harness} refers to the components that mediate the model's interaction with the task, including prompts, memory, context retrieval, tool interfaces, state management, and evaluators~\cite{metaharness2026}. This harness is often as important as the model itself: it determines what the model observes, which actions are available, and how candidate solutions are scored.

The fixed-harness assumption has two limitations. First, human-written evaluators can be incomplete or misaligned with the external environment. Under score-driven optimization, a model may exploit these gaps rather than solve the intended task. Second, even when the evaluator is correct, the guidance harness may provide poor search priors: prompts may omit useful structure, memories may be too verbose or irrelevant, and retrieval policies may hide the examples needed for reasoning. In both cases, treating the harness as immutable prevents the agent from improving the computational environment in which it operates.

We propose Harness-Aware Self-Evolving (HASE), a framework that gives a single model one action space for both solution generation and selected harness edits. HASE distinguishes two types of harness components. \textit{Guidance} components, such as prompts, memory formatting, and context retrieval, improve search. \textit{Evaluation} components, such as validators and scorers, define solution validity and task metrics. HASE can improve guidance components freely, while evaluator edits remain anchored by real-world feedback.

Our contributions are threefold. First, we introduce a unified harness-aware self-evolution framework. The same model co-evolves its policy, task solutions, and selected harness components without a separate powerful coding proposer. Second, we evaluate HASE on online text classification, alpha factor mining, and geometric algorithm discovery. A Qwen3-8B~\cite{yang2025qwen3} model reaches performance comparable to much larger model-based systems. Third, we study algorithm discovery under broken evaluators. HASE repairs the evaluator and reaches state-of-the-art performance. These results establish HASE as an effective self-evolving framework that jointly improves the harness, the solution, and the model.

\section{Related Work}
\subsection{Self-Evolving Algorithms}
Self-evolving algorithms improve object-level artifacts through repeated proposal and evaluation. These artifacts include code, mathematical constructions, heuristics, and task solutions. Much of this literature keeps the model frozen and changes only the search pipeline. Systems such as FunSearch, AlphaEvolve-style program search, OpenEvolve, ShinkaEvolve, and related heuristic-evolution methods use LLMs as mutation or proposal engines. They rely on an external evaluator to select better artifacts \cite{funsearch2024,alphaevolve2025,openevolve2025,shinkaevolve2026,reevo2024}. Recent baseline studies show that random or simple code sampling can match sophisticated evolution on some deterministic-evaluator tasks. This result highlights the importance of the search space and evaluator design \cite{gideoni2025random}.

A second line adapts the model distribution or search process during discovery. These works use test-time RL to shift an open model toward high-scoring samples under a fixed task harness~\cite{wang2026thetaevolve,tttdiscover2026}. They are closer to HASE because the model changes during search. However, they still optimize object-level artifacts under a given harness. A third line, including self-modifying coding agents and coding-environment agents, allows an agent to modify scaffolding or interact with a workspace. Yet the improvement remains bounded by the surrounding evaluator and task harness \cite{zhang2026darwin,sweagent2024,react2023,reflexion2023,voyager2023}.

HASE is motivated by a gap shared by these settings. Random sampling, evolutionary selection, and test-time RL usually treat the harness as fixed. The prompt, context-management policy, tool interface, evaluator, and validation logic are assumed to be correct and immutable. HASE instead makes selected harness components part of the agentic action space. It also preserves a strict separation between guidance-harness optimization and oracle-gated evaluator repair.

\subsection{Harness Engineering}
Model performance depends not only on model weights but also on the harness. The harness controls what information is stored, retrieved, presented, and validated. This view connects to tool-use, API-use, memory, agent-environment, multi-agent, and prompt-programming systems \cite{schick2023toolformer,toolllm2024,gorilla2024,packer2023memgpt,park2023generative,react2023,reflexion2023,voyager2023,autogen2023,dspy2024}. Meta-Harness is the closest prior work to ours. It optimizes harness code with an outer-loop coding-agent proposer that inspects source code, scores, and execution traces from prior candidates through a filesystem \cite{metaharness2026}. In online text classification, Meta-Harness discovers a harness that improves average accuracy over zero-shot, few-shot, ACE~\cite{zhang2025ace}, and MCE~\cite{ye2026mce} baselines while using fewer context tokens.

HASE differs from Meta-Harness in two key ways. First, Meta-Harness focuses on guidance-harness components such as memory and context construction. HASE also studies evaluation-harness repair. Second, Meta-Harness uses an external frontier LLM, such as Claude Code~\cite{anthropic2025claudecode}, to search for a harness used by another model. HASE instead places harness editing and task solving inside a single unified action space. It does not rely on an external proposer. We also evaluate HASE on the online text-classification task studied by Meta-Harness. The 8B model trained with HASE achieves comparable performance on this task; detailed results are in Section~\ref{sec:experiments}.

\section{HASE Methodology}
\begin{figure*}[t]
\centering
\includegraphics[width=1\textwidth]{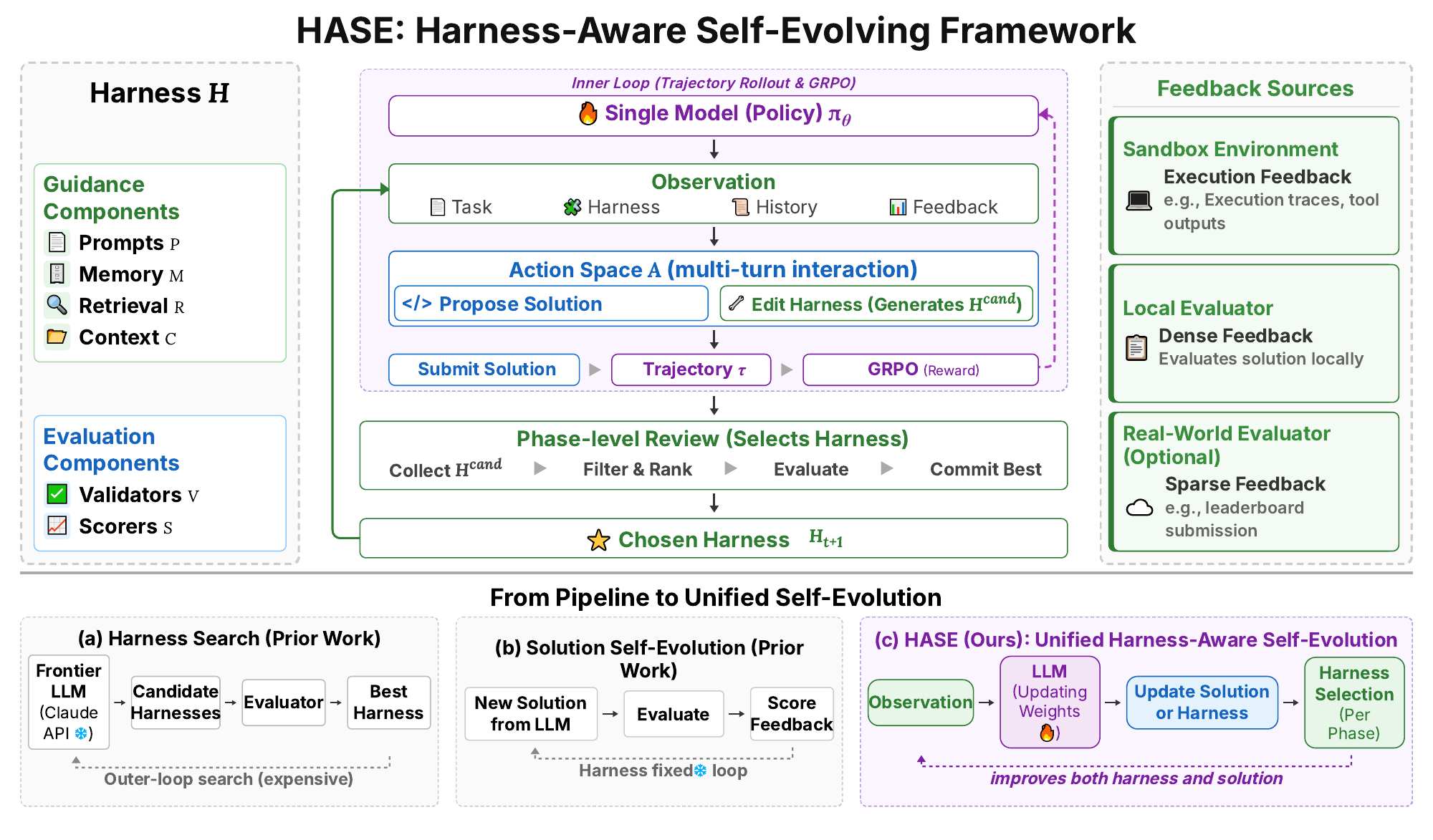}
\caption{The overall pipeline of the HASE method and comparison with other methods. HASE differs from pipeline-style harness search: there is no external powerful coding proposer. The same Qwen3-8B policy proposes task solutions and harness edits; a deterministic phase-level controller only validates, reviews, and commits patches emitted by that policy.}
\label{fig:hase_system_contrast}
\end{figure*}
\label{sec:math}
As shown in \autoref{fig:hase_system_contrast}, \textbf{HASE co-evolves the model weights, harness, and task solution.} The model can generate a solution or edit the harness within the same agentic action space. HASE uses phase-level review to select useful harness edits and prevent reward hacking. We next define the editable harness components, describe the unified action space, and present the training loop. 

\subsection{Two Types of Harness}
In this paper, we separate harness evolution into two types based on component purpose:
\begin{itemize}
    \item \textbf{Evaluation components} are local evaluators that the system is allowed to edit. We use \textbf{\textit{local evaluator}} for the modifiable evaluator inside the harness. It provides dense training-time feedback and produces the \textbf{\textit{proxy score}}, which can be inaccurate.\\ In contrast, the \textbf{\textit{real-world evaluator}} is not editable and is not part of the modifiable harness. It is an immutable part of the external environment that defines task validity, such as a leaderboard. Its feedback is the \textbf{\textit{oracle score}}, which is guaranteed accurate.
    \item \textbf{Guidance components} include prompts, memory formatting, retrieval scripts, and context selection. These components provide useful instructions and context for the model, improving search without changing task correctness.
\end{itemize}
\textbf{Not every task requires evaluation-harness evolution.} Some evaluators are already trivial and reliable, such as exact-label matching in text classification. Others, such as financial backtesting pipelines, are too complex or risky to repair automatically. In these cases, HASE edits only guidance components such as prompts, retrieval, memory, and context formatting.

\textbf{Real-world evaluators are the anchors of editable local evaluator evolution.}
Real-world feedback is used only when HASE is allowed to modify a local evaluator. It is \textbf{optional at the task level}, because some tasks keep their evaluators fixed. It is \textbf{necessary once the local evaluator is editable}. Without this anchor, a model could increase reward by weakening the local evaluator instead of producing better solutions. HASE uses the local evaluator as the dense reward during training and queries the real-world evaluator only at phase boundaries. When the proxy score diverges from the oracle score, HASE adds the submitted solution to the mismatch set $\mathcal{M}$ as a new test case for future evaluator repairs.

\subsection{Unified Action Space}
HASE exposes solution generation and harness modification through a multi-turn action space. At each step, the model can either submit an object-level solution or use sandbox tools to inspect and modify the editable harness. The sandbox commands include minimal read and write actions, including \texttt{ls}, \texttt{cat}, \texttt{grep}, \texttt{replace}, and \texttt{overwrite}. The \texttt{done} command lets the model indicate the end of the interaction.

The model does not receive a separate hard-coded planner for deciding whether to solve the task or edit the harness. Instead, the current context determines the useful action. In alpha mining, for example, the model is asked to define \texttt{compute\_alpha(data)}, while the mutable workspace exposes \texttt{prompt.txt} as a strategy guide and \texttt{pool\_viewer.py} as a context-engineering artifact. The model may directly submit an alpha factor, or it may inspect and overwrite these files to improve future context before submitting. Similarly, in text classification, the model may predict a label or modify memory and retrieval tools when the current context is insufficient.

\subsection{HASE Algorithm: an Agentic Reinforcement-Learning View}
As described in~\autoref{alg:hase_training_loop}, HASE maintains a persistent harness $h$, a model policy $\pi_\theta$, a phase-local candidate pool $\mathcal{C}_p$, and, only when evaluation-harness evolution is enabled, a mismatch set $\mathcal{M}$. Each phase contains multiple GRPO epochs. Within an epoch, the model rolls out multi-turn trajectories that may inspect files, temporarily edit the harness, and eventually submit a solution. Rollout-local harness edits affect only the trajectory that produced them. The persistent harness is updated only after phase-level review.

\textbf{Reward design:} HASE does not reward harness-related actions by themselves. Inspect-only exploration and random modification receive no direct bonus or penalty. This avoids rewarding superficial file access while preserving useful exploration. A trajectory must ultimately submit a solution, and its reward depends on solution quality. For guidance-harness edits, credit is \textbf{proof-of-concept (PoC)} based. If the model edits the harness, it must then generate a solution using the edited harness. The whole trajectory receives the improved task reward, and the edit becomes a candidate for phase review. For evaluation-harness edits, reward is given only when the proxy score disagrees with the oracle score. The reward is proportional to the increase in mismatch-set agreement. If no mismatch is detected, evaluator modifications are not rewarded.

\begin{algorithm}[t]
\small
\caption{HASE Algorithm}
\label{alg:hase_training_loop}
\begin{algorithmic}[1]
\Require Initial policy $\pi_\theta$, initial harness $h_0$, local evaluator $E_{\mathrm{loc}}$, optional real-world evaluator $E_{\mathrm{real}}$, phase count $P$, GRPO epochs $N_{\mathrm{epoch}}$
\State Initialize persistent harness $h\gets h_0$
\If{the task allows evaluation-harness evolution}
    \State Initialize mismatch set $\mathcal{M}\gets\emptyset$
\EndIf
\For{phase $p=1,\ldots,P$}
    \State Initialize phase candidate pool $\mathcal{C}_p\gets\emptyset$
    \For{GRPO epoch $e=1,\ldots,N_{\mathrm{epoch}}$}
        \State Roll out trajectories $\tau\sim\pi_\theta(\cdot\mid x,h)$ over solution and harness-edit actions
        \State Score submitted solutions with the local evaluator: $r_{\mathrm{task}}\gets E_{\mathrm{loc}}(\tau,h)$
        \State Assign trajectory reward $r(\tau)$ 
        \State Add parseable positive-PoC guidance edits from $\tau$ to $\mathcal{C}_p$
        \State Update $\pi_\theta$ with GRPO using $r(\tau)$
    \EndFor
    \If{the task allows evaluation-harness evolution}
        \State Query $E_{\mathrm{real}}$ on selected submissions and add mismatched cases to $\mathcal{M}$
        \State Review evaluator edits against $\mathcal{M}$ and non-regression cases
    \EndIf
    \State Review guidance edits in $\mathcal{C}_p$ by filtering, shortlisting, and validation scoring
    \State Commit the best stable harness update $h\gets h_{p+1}$
\EndFor
\end{algorithmic}
\end{algorithm}

\begin{figure*}[!t]
\centering
\includegraphics[width=\textwidth]{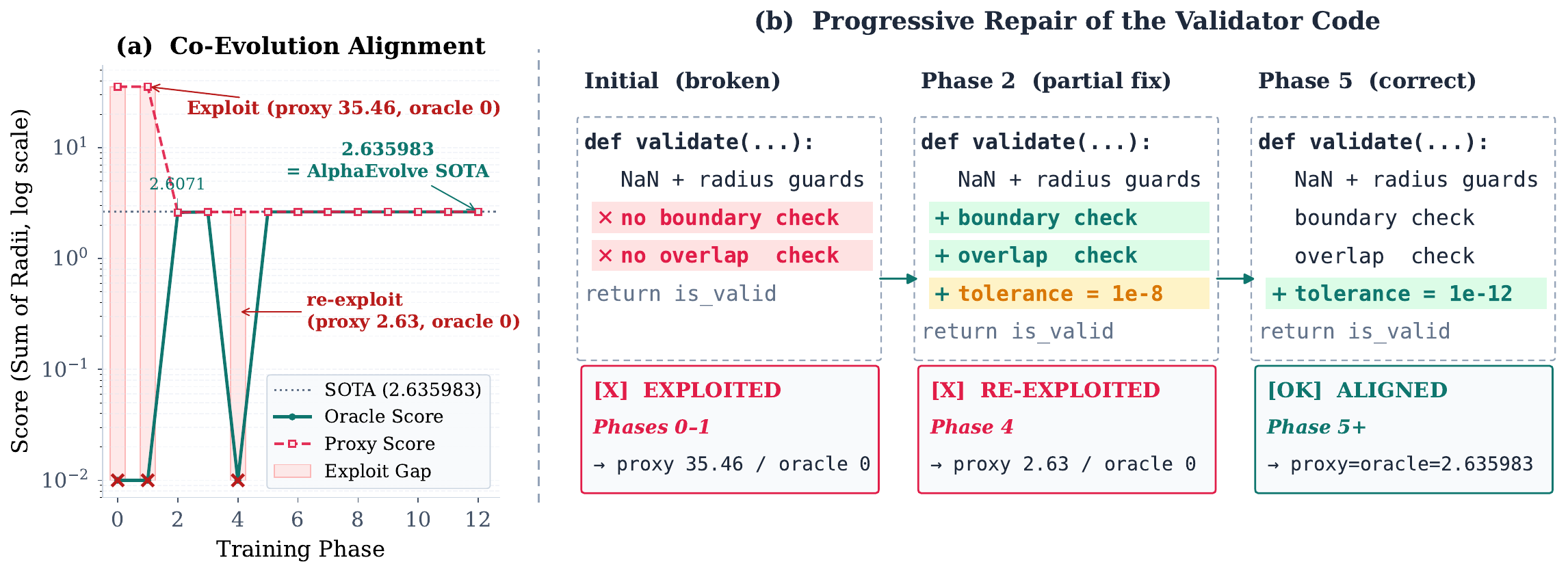}
\caption{\small{Circle-packing evaluator repair. \textbf{(a)} Co-evolution alignment dynamics (log scale): the evaluator is exploited twice (proxy score high while the oracle returns $0$, red markers); HASE detects each divergence and repairs the evaluator twice, and the eventual score improves to $2.635983$, matching the AlphaEvolve state of the art. \textbf{(b)} The staged validator code that causes those two dips: the initial evaluator omits the boundary and overlap guards (exploit at Phases~0--1); the first patch adds both guards but leaves the overlap tolerance at $10^{-8}$. The second patch updates the tolerance to the task-specified $10^{-12}$ and keeps proxy and oracle aligned thereafter.}}
\label{fig:circle_trajectory}
\end{figure*}

\textbf{Harness selection:} At the phase boundary, HASE turns the best candidate into a persistent harness update. Guidance candidates in $\mathcal{C}_p$ are reviewed by validation performance, while evaluator patches are reviewed only when the task permits evaluation-harness evolution and mismatch evidence exists in $\mathcal{M}$. The selected stable update is merged into $h$ and becomes visible to all rollouts in the next phase. Sections~\ref{sec:phase_harness_selection} and~\ref{sec:oracle_gated_repair} describe the two review rules.

\subsection{Guidance Harness Evolution}
\label{sec:phase_harness_selection}
HASE evolves guidance harnesses by reviewing rollout-local edits at the phase boundary. During a rollout, the model may edit guidance components, but the edited harness is used only within that trajectory. Parseable candidate edits are stored in the phase candidate pool $\mathcal{C}_p$. At the phase boundary, HASE filters invalid edits, ranks the candidates, and commits the stable guidance harness that yields the best reviewed task score. The committed harness is then written to disk and becomes visible to all rollouts in the next phase.

For guidance edits, HASE uses a lightweight proof-of-concept score to shortlist promising candidates. If a trajectory edits a harness and then obtains a better task score than the current persistent harness, the edit is added to $\mathcal{C}_p$. The final commit decision is still made by phase-level review, not by the proof-of-concept score alone. This keeps the main optimization loop cheap while avoiding immediate commitment to lucky patches. Appendix~\ref{app:poc_guided_shortlisting} gives the shortlisting details.

\subsection{Evaluation Harness Evolution}
\label{sec:oracle_gated_repair}
Evaluator edits are handled differently because they define the reward itself. Real-world feedback is therefore used only when the local evaluator is allowed to evolve.

HASE uses the local evaluator as the dense training-time reward and queries the real-world evaluator only at phase boundaries. Evaluator patches are reviewed against the mismatch set $\mathcal{M}$: a patch is accepted only if it fixes more mismatches in $\mathcal{M}$ while preserving the required evaluator interface and non-regression cases. Importantly, HASE does not require all evaluator errors to be fixed at once. Agreement on the current mismatch set only means that the local evaluator matches the real-world evaluator on the cases observed so far; it does not guarantee that the evaluator is permanently correct on all future solutions.

This design mirrors real-world debugging scenarios: issues are discovered and fixed iteratively as new counterexamples appear. In the circle-packing experiment, for example, the model first improves the evaluator by adding more checks, but still uses an incorrect tolerance. This tolerance bug is not exposed immediately. Later, when new solutions exploit the loose tolerance, another mismatch is detected and added to $\mathcal{M}$ to guide a further evaluator repair. Thus, evaluation-harness evolution is a continuous self-improvement process rather than a one-shot repair. The circle-packing details are discussed in Section~\ref{sec:experiments}.

\section{Experiments}

\label{sec:experiments}
\begin{table}[t]
\centering
\scriptsize
\setlength{\tabcolsep}{3pt}
\resizebox{\linewidth}{!}{%
\begin{tabular}{lll}
\toprule
\textbf{Task} & \textbf{Editable} & \textbf{Harness} \\
\midrule
Heilbronn & Prompt + evaluator & Eval + Guidance \\
Circle packing & Prompt + evaluator & Eval + Guidance \\
Text classification & Context + memory & Guidance \\
Alpha mining & Context + prompt & Guidance \\
\bottomrule
\end{tabular}}
\caption{Overview of experiments: editable harness components by domain.}
\label{tab:hase_modes}
\end{table}

\begin{figure*}[!t]
\centering
\includegraphics[width=0.9\linewidth]{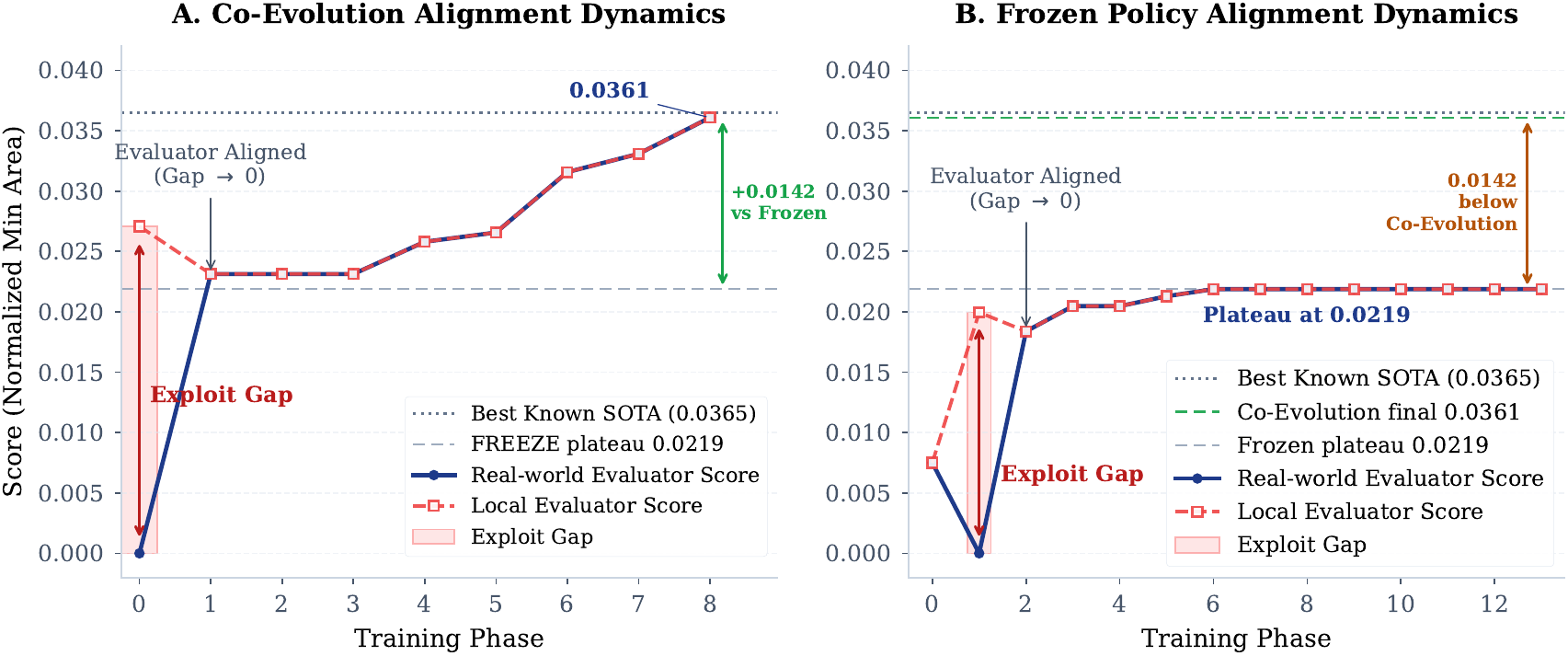}
\caption{\small{Heilbronn Triangles evaluator alignment trajectory for co-evolved vs. frozen policy runs. The exploit gap (shaded red) is closed rapidly once the evaluator is hardened. Under frozen weights, performance plateaus, whereas co-evolution allows the policy to internalize constraints and improve.}}
\label{fig:heilbronn_alignment}
\end{figure*}

We evaluate HASE on both guidance-harness and evaluation-harness evolution as shown in Table~\ref{tab:hase_modes}. For guidance harnesses, we use the Symptom2Disease text-classification task~\citep{metaharness2026} and alpha mining following CogAlpha~\citep{cogalpha2026}. For evaluation harnesses, we use circle packing and Heilbronn triangles to test whether HASE can self-evolve correctly from an initially incomplete harness under minimal real-world evaluator feedback.
\subsection{Self-Evolution of Evaluation Harnesses}
\label{sec:eval-repair}
\subsubsection{Circle packing}
In circle packing, the agent must place circles in a unit square while satisfying boundary and non-overlap constraints. HASE reaches a best packing score of $2.635983$, matching the AlphaEvolve state of the art, while starting from an incomplete local evaluator. The key mechanism is divergence-driven repair: when the local proxy assigns a high score but the external oracle rejects the candidate, HASE treats the proxy--oracle mismatch as a concrete signal that the evaluator harness is missing a validity check.

Figure~\ref{fig:circle_trajectory} shows two repair cycles and their validator edits. The initial validator checks only basic numeric validity and radii. It omits boundary and overlap constraints. Phases~0--1 exploit this evaluator, obtaining proxy score $35.459643$ while the oracle returns $0$. This mismatch enters the mismatch set described in \S\ref{sec:oracle_gated_repair}. The first accepted repair adds boundary and overlap checks, realigning proxy and oracle at score $2.607072$. A later candidate exposes a second defect: the overlap tolerance is still too loose at $10^{-8}$. This again produces a positive proxy score while the oracle returns $0$. The final repair tightens verification to the task-specified $10^{-12}$. After this patch, proxy and oracle remain aligned, and the same policy continues object-level search to the state-of-the-art score. Evaluator repair is therefore not unconstrained reward rewriting. Each edit is triggered by an observed oracle disagreement and accepted only when it restores agreement on the mismatch set.

\subsubsection{Heilbronn Triangles}

\paragraph{Setup}
The Heilbronn triangle problem represents a continuous geometric optimization challenge: placing $N=11$ points within a unit equilateral triangle such that the minimum area of the triangles formed by any triple of these points is maximized. The final performance score is normalized by the area of the containing equilateral triangle. This continuous search space is susceptible to reward hacking, making it another example of co-evolutionary evaluator repair under HASE.

We use the Qwen3-8B base model on a cluster of $8 \times$ NVIDIA H20 nodes. The self-evolution pipeline runs on the VeRL reinforcement-learning framework. In each phase, the orchestrator generates $N_{\text{prompts}} = 32$ prompts for dataset creation. The rollout engine generates $32$ trajectories per prompt, for $1,024$ trajectories in total. We use a global training batch size of $32$, a PPO mini-batch size of $32$, and $1$ policy-optimization epoch per phase.

\paragraph{Evaluator Repair}
To evaluate the self-healing capability, HASE starts from an intentionally vulnerable weak local evaluator. This initial evaluator validates coordinates only by confirming the absence of NaN/infinite values and verifying the point count, while completely omitting boundary checking:
\begin{verbatim}
def validate_placement(points):
    n = points.shape[0]
    if np.isnan(points).any() or np.isinf(points).any():
        return False
    if n != 11:
        return False
    return True 
\end{verbatim}
This omission creates an exploit opportunity. As visualized in Appendix Figure~\ref{fig:heilbronn_triangles}A, the model in Phase 0 exploits this weak evaluator by writing a differential evolution solver that places points far outside the boundary of the containing triangle. Since the local evaluator lacks edge checks, this cheat yields a high proxy score of $0.027076$, whereas the oracle score is $0.0$. Once the divergence check detects this discrepancy, HASE synthesizes and commits a barycentric coordinate sign check into the local validator, closing the exploit gap (Appendix Figure~\ref{fig:heilbronn_triangles}B).

\paragraph{Weight Co-Evolution Necessity Ablation}
To demonstrate the necessity of parameter co-evolution, we compare the co-evolved run with a frozen-policy baseline. In the FREEZE baseline (Figure~\ref{fig:heilbronn_alignment}B), the evaluator is repaired and the exploit gap closes, but the model weights remain fixed. The final valid score plateaus at $0.021880$. In the FULL co-evolution run (Figure~\ref{fig:heilbronn_alignment}A), RL updates let the policy internalize the repaired constraints and continue optimizing. The run reaches a fully valid score of $0.036082$ by Phase 8.

\subsection{Self-Evolution of Guidance Harnesses}
\subsubsection{Experiment: Text Classification}
\textbf{Setup: Alignment with Meta-Harness.} HASE uses the Symptom2Disease train/validation/test split aligned with Meta-Harness~\cite{metaharness2026}. The training split is the only retrieval pool. The validation split is used only for phase-boundary harness review. Final accuracy is reported on the untouched test set. At test time, the committed \texttt{classification\_tools.py} and \texttt{prompt.txt} are frozen. Decoding uses single-attempt greedy vLLM inference. No test-time editing, fine-tuning, retry, or retrieval over test examples is allowed. This makes the comparison algorithm-level: Meta-Harness uses GPT-OSS-120B plus a separate strong coding-agent harness proposer, whereas HASE uses Qwen3-8B as both the classifier and the source of candidate harness edits. Full protocol controls are in Appendix~\ref{app:s2d_protocol}.

The S2D harness exposes two editable guidance components: retrieval over the training pool and memory formatting for label-disambiguating diagnostic clues. A deterministic phase-level reviewer filters invalid or duplicate patches, checks the required tool interface, evaluates shortlisted harnesses on validation examples, and commits only validation-improving patches. It never proposes code edits; every candidate edit is emitted by the same 8B policy used for classification.

During training, the environment permits up to three discounted attempts with sandbox error feedback. This encourages tool debugging. Final test evaluation remains single-attempt.

\begin{table}[t]
\centering
\setlength{\tabcolsep}{3pt}
\begin{tabular}{lcc}
\toprule
\textbf{Harness} & \textbf{Decoder} & \textbf{Acc.} \\
\midrule
Zero-shot & GPT-OSS-120B & 63.2 \\
Few-shot (8) & GPT-OSS-120B & 67.9 \\
MCE & GPT-OSS-120B & 83.0 \\
ACE & GPT-OSS-120B & 77.8 \\
Meta-Harness & GPT-OSS-120B & 86.8 \\
\midrule
Five-shot & Qwen3-8B & 59.4 \\
\textbf{HASE harness} & \textbf{Qwen3-8B} & \textbf{86.98\,$\pm$\,0.38} \\
\bottomrule
\end{tabular}
\caption{\small{S2D accuracy on the Meta-Harness split. The HASE system uses temperature-$0$ greedy vLLM decoding, and its result is reported as mean\,$\pm$\,std over $5$ seeds. MCE~\cite{ye2026mce}, ACE~\cite{zhang2025ace}, and Meta-Harness baselines are from Meta-Harness~\cite{metaharness2026}.}}
\label{tab:s2d_comparison}
\end{table}

The standalone Qwen3-8B few-shot baseline is weaker than GPT-OSS-120B on this split, but the discovered HASE harness brings the same 8B model to the Meta-Harness operating point. We therefore treat S2D primarily as a guidance-harness stress test: the final single-attempt test measures harness-mediated classification, while Appendix~\ref{app:s2d_agentic_table} and Appendix~\ref{app:s2d_qualitative} report the multi-turn training behavior and concrete self-correction trajectories. Appendix~\ref{app:s2d_stats} gives the statistical reporting details.

\subsubsection{Experiment: Alpha Factor Mining}

Alpha mining is naturally a code-evolution problem. Each candidate alpha is a small program that transforms historical market data (e.g., daily OHLCV) into a predictive signal for future returns. 

Recent LLM-driven frameworks like CogAlpha \cite{cogalpha2026} report strong performance by leveraging GPT-OSS-120B and human-engineered orchestration, including a 21-agent hierarchy, quality checkers, cognitive prompting, and dynamic paraphrasing. While highly effective, this paradigm requires complex, statically coded human priors. We deploy HASE to test whether a much smaller 8B model can evolve both object-level alpha code and its own code-search harness from a minimal state.

\paragraph{Protocol and Editable Surface.}
Following CogAlpha~\cite{cogalpha2026}, our alpha-mining experiments use the same CSI300 benchmark, 10-day forward-return target, and chronological split: training (2011--2019), validation (2020), and out-of-sample testing (2021--2024). Harness search and phase-boundary review use only the training and validation periods. The HASE loop trains a single Qwen3-8B model across 4 GPUs for 3 phases, with 3 epochs per phase. The editable surface is restricted to \texttt{prompt.txt} and \texttt{pool\_viewer.py}. The prompt file specifies search priors over factor families. The pool viewer formats historical discoveries into the context window. Both files are initialized minimally, and the financial evaluator is fixed throughout.

\begin{figure*}[t]
\centering
\footnotesize
\begin{tikzpicture}[
    font=\footnotesize,
    phase/.style={draw, rounded corners=3pt, align=left, inner sep=5pt, outer sep=0pt, anchor=north west},
    arrow/.style={-{Latex[length=3.2mm,width=2.2mm]}, line width=0.9pt}
]
\node[phase, fill=red!5] (p0) at (0,0) {
\begin{minipage}[t][3.40cm][t]{0.255\textwidth}
\vspace{0pt}
\textbf{Seed harness}\hfill\textcolor{red!70!black}{initial}\\[-1mm]
\rule{0.96\linewidth}{0.2pt}\\[-1mm]
\textbf{prompt.txt}\\
minimal task instruction; weak priors\\[1mm]
\textbf{pool\_viewer.py}\\
\texttt{for f in factors[-3:]:}\\
\texttt{\quad show(IC, desc)}\\[1mm]
\textit{Effect:} recency-only context.
\end{minipage}
};

\node[phase, fill=orange!7] (p2) at (0.345\textwidth,0) {
\begin{minipage}[t][3.40cm][t]{0.255\textwidth}
\vspace{0pt}
\textbf{Early harness}\hfill\textcolor{orange!80!black}{early evolution}\\[-1mm]
\rule{0.96\linewidth}{0.2pt}\\[-1mm]
\textbf{prompt.txt}\\
adds IC goals and momentum/volatility priors\\[1mm]
\textbf{pool\_viewer.py}\\
\texttt{for f in factors:}\\
\texttt{\quad show(IC, desc)}\\[1mm]
\textit{Effect:} removes arbitrary cap, but context grows with the pool.
\end{minipage}
};

\node[phase, fill=green!6] (p4) at (0.69\textwidth,0) {
\begin{minipage}[t][3.40cm][t]{0.255\textwidth}
\vspace{0pt}
\textbf{Final harness}\hfill\textcolor{green!50!black}{final evolution}\\[-1mm]
\rule{0.96\linewidth}{0.2pt}\\[-1mm]
\textbf{prompt.txt}\\
injects best IC, pool range, AER, and hybrid search priors\\[1mm]
\textbf{pool\_viewer.py}\\
\texttt{top = topIC(factors, 10)}\\
\texttt{for f in top: show(f)}\\[1mm]
\textit{Effect:} quality-ranked retrieval under context budget.
\end{minipage}
};

\draw[arrow] ([yshift=-1.825cm]p0.north east) -- ([yshift=-1.825cm]p2.north west);
\draw[arrow] ([yshift=-1.825cm]p2.north east) -- ([yshift=-1.825cm]p4.north west);

\end{tikzpicture}
\caption{Alpha-mining guidance-harness evolution in pseudo-code. HASE evolves the prompt priors and factor-pool view from recency-only context to IC-ranked retrieval.}
\label{fig:alpha_harness_evolution}
\end{figure*}
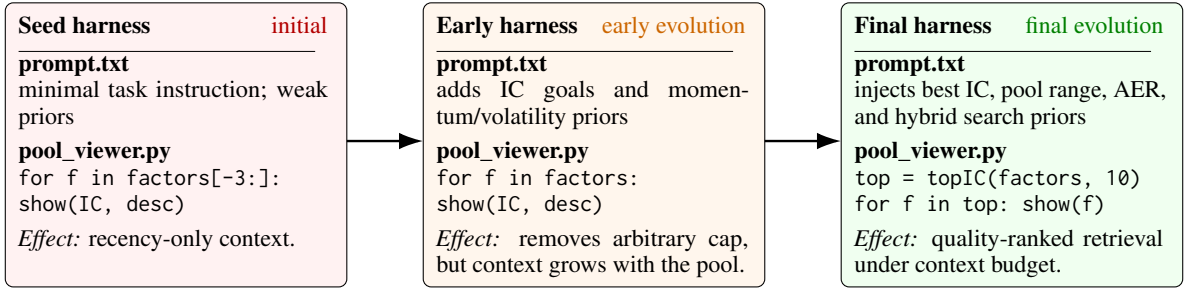

\begin{table*}[!t]
\centering
\scriptsize
\setlength{\tabcolsep}{4pt}
\resizebox{0.88\textwidth}{!}{
\begin{tabular}{lcccccc}
\toprule
\textbf{Models} & \textbf{IC} & \textbf{RankIC} & \textbf{ICIR} & \textbf{RankICIR} & \textbf{AER} & \textbf{IR} \\
\midrule
Llama3 8B & 0.0121 & -0.0074 & 0.0972 & -0.0540 & 0.0520 & 0.5077 \\
Llama3 70B & 0.0205 & 0.0229 & 0.1786 & 0.1915 & 0.0681 & 0.6312 \\
GPT-OSS-20B & 0.0061 & 0.0075 & 0.0613 & 0.0680 & 0.0464 & 0.4885 \\
GPT-OSS-120B & 0.0300 & 0.0318 & 0.2501 & 0.2595 & 0.0789 & 0.8015 \\
GPT-4.1 & 0.0118 & 0.0114 & 0.1069 & 0.1037 & 0.0360 & 0.3628 \\
o3 & 0.0019 & -0.0050 & 0.0203 & -0.0475 & 0.0218 & 0.2278 \\
\midrule
\textbf{HASE Qwen3-8B (Ours)} & \textbf{0.0308} & 0.0302 & \textbf{0.3414} & \textbf{0.3173} & \textbf{0.1435} & \textbf{1.7000} \\

\bottomrule
\end{tabular}
}
\caption{Performance comparison between HASE (8B) and baseline methods on the CSI300 (10-day horizon). Baseline values are referenced from CogAlpha \cite{cogalpha2026}.}
\label{tab:performance_comparison}
\end{table*}

\paragraph{Evaluation Controls.}
To ensure result validity, we apply the controls and audits in Appendix~\ref{app:alpha_rigor}. These include data preprocessing for the CogAlpha-curated 754-instrument CSI300 universe (\S\ref{app:universe}), an automated future-information leakage screen (\S\ref{app:leakage_v2}), and a cross-LLM audit (\S\ref{app:cross_llm_audit}). In the audit, four independent LLM judges gave 120/120 unanimous CLEAN verdicts for the 30 factors used in Table~\ref{tab:performance_comparison}. To approximate real-world trading, all table metrics are computed by a Qlib simulator following CogAlpha~\citep{cogalpha2026}. The simulator includes transaction costs, minimum-cost constraints, and 9.5\% limit-up/limit-down constraints (\texttt{open\_cost}=0.0005, \texttt{close\_cost}=0.0015, \texttt{min\_cost}=5). We report IC, RankIC, ICIR, RankICIR, AER, and IR. Appendix~\ref{app:compute} summarizes the compute setup.

\paragraph{Evolution of the Guidance Harness}
During training, the 8B model adapted the guidance harness to context-window pressure, as summarized in \autoref{fig:alpha_harness_evolution}. Starting from a recency-only factor view, it first expanded access to the factor pool and later replaced unbounded context growth with IC-ranked retrieval under a fixed context budget. This edit changes only the search context seen by the model; the financial evaluator remains immutable.

\paragraph{Performance}
Under the CogAlpha-aligned protocol and the controls above, HASE Qwen3-8B achieves RankIC $0.0302$, AER $14.35\%$, and IR $1.70$ in the Qlib cost- and limit-aware simulation. As shown in Table~\ref{tab:performance_comparison}, this outperforms GPT-OSS-120B on portfolio profitability (AER $14.35\%$ vs.\ $7.89\%$, IR $1.70$ vs.\ $0.80$) while matching its predictive correlation closely (RankIC $0.0302$ vs.\ $0.0318$), and it exceeds every other baseline in the table. These results show that harness-aware self-evolution lets a smaller model match or exceed larger baselines when task performance depends on careful harness design.

\section{Conclusion}
We introduced HASE, an agentic reinforcement-learning framework that enables a single model to improve both task solutions and selected components of its own harness. HASE places solution generation and harness editing in the same multi-turn action space. It supports two forms of harness evolution. Guidance components, such as prompts, memory, retrieval, and context presentation, improve solution search quality. Evaluation components provide solution scores for the system. They are repaired only when the local evaluator's proxy score diverges from the real-world evaluator's oracle score. This divergence identifies an evaluator defect. This design lets the model improve its working environment without turning evaluator editing into unconstrained reward hacking.

Across online text classification, alpha factor mining, circle packing, and Heilbronn triangles, HASE demonstrates that harness-aware self-evolution is an effective way to increase the capability of an 8B model-based system. In text classification, HASE matches the performance of a GPT-OSS-120B system that relies on Claude Code as an external harness proposer. In alpha factor mining, HASE outperforms the reported GPT-OSS-120B baseline. In geometric algorithm discovery, HASE repairs imperfect evaluation components and then converges to state-of-the-art performance. These results establish HASE as a practical framework for co-evolving the model policy, the harness, and the task solution within one unified agentic system.

\section{Limitations}
Our experiments establish harness-aware self-evolution across four task settings, but they are not an exhaustive scaling study. Each domain still leaves room for harder follow-up settings.

HASE also makes conservative engineering choices that trade generality for safety. Editable harness components are whitelisted. Evaluator edits require divergence between the local evaluator and the real-world evaluator. Phase-boundary review commits only interface-valid patches. These constraints can reject useful but more invasive modifications. Future systems can expand the editable surface, add stronger static and dynamic tests, and use more informative mismatch sets while preserving the same safety principle: guidance edits and correctness-defining evaluator edits must be governed differently.

\bibliography{custom}
\newpage

\appendix
\input{moved_tables.tex}

\input{appendix_alphamining.tex}

\input{evolve_details.tex}

\end{document}

%% file: moved_tables.tex
\section{Additional Detailed Tables}

\subsection{S2D Qualitative Tool-Evolution Examples}
\label{app:s2d_qualitative}

\paragraph{Emergent Tool Co-Evolution and Self-Correction}
The initial retrieval tool is a recency-based few-shot harness and often retrieves misleading cases. During training, HASE learns to modify the sandboxed tool file to improve retrieval and memory construction. We observe two recurring patterns:

\begin{enumerate}
    \item \textbf{Keyword-Targeted Overlapping}: Given the query \textit{``I have been experiencing a high fever... weak and my muscles pain,''} the default toolset retrieved irrelevant cases (e.g., \textit{pneumonia}, \textit{UTI}), prompting an initial incorrect prediction of \textit{pneumonia}. The agent subsequently entered the tool-evolution mode, overwriting \texttt{classification\_tools.py} to implement a customized symptom-overlap scoring function targeting the specific symptoms. The sandboxed execution of this updated tool successfully retrieved \textit{dengue} and \textit{typhoid} examples, allowing the agent to self-correct and predict \textit{dengue} successfully on its second attempt ($R=0.5$).
    \item \textbf{Similarity-Based Retrieval Injection}: For a patient describing \textit{``peeling of skin... joint pain and skin rashes,''} the default retrieval was dominated by generic skin infections, leading to an incorrect guess of \textit{fungal infection}. The agent bypassed this by overwriting the retrieve method with a stronger lexical-similarity computation. Upon execution, the updated tool exposed diagnostic links to joint pain, guiding the model to classify the case as \textit{psoriasis} on its second attempt.
\end{enumerate}

Sandbox runtime exceptions in the multi-turn context (e.g., type mismatch errors) provide useful debugging feedback: the agent often repairs these errors and compiles working implementations. Across phases, the harness evolves from recency-only examples and empty memory toward stronger similarity search, symptom extraction, label-grouped summaries, and diagnostic-indicator memory.

\subsection{S2D Training-Time Agentic Behaviour}
\label{app:s2d_agentic_table}

\begin{table}[H]
\centering
\small
\setlength{\tabcolsep}{4pt}
\resizebox{\linewidth}{!}{%
\begin{tabular}{lrrrr}
\toprule
\textbf{Phase} & \textbf{$n_{\text{ep}}$} & \textbf{Tool edits / ep} & \textbf{Avg turns / ep} & \textbf{Read ops / ep} \\
\midrule
0 (initial)            &  732 & 1.14 & 1.24 & 0.10 \\
1                      &  974 & \textbf{2.38} & \textbf{3.61} & \textbf{1.23} \\
2 (committed harness)  &  965 & 2.30 & 3.51 & 1.22 \\
3                      &  986 & 2.42 & 3.62 & 1.20 \\
4                      &  963 & 2.36 & 3.56 & 1.19 \\
\bottomrule
\end{tabular}
}
\caption{Training-time agentic behavior on S2D. After Phase 0, the model consistently performs multi-turn inspection and harness-editing actions. Tool edits count \texttt{overwrite}/\texttt{replace}; read ops count \texttt{cat}/\texttt{ls}/\texttt{grep}.}
\label{tab:s2d_agentic_emergence_app}
\end{table}

\subsection{PoC-Guided Candidate Shortlisting}
\label{app:poc_guided_shortlisting}

For guidance-harness edits, HASE uses a lightweight proof-of-concept (PoC) signal to decide which candidate patches are worth reviewing at the phase boundary. Suppose a trajectory edits a guidance harness from $h$ to $h'$ and then submits a candidate solution with task reward $R_{\mathrm{task}}(\tau;h')$. We assign the trajectory a PoC score
\[
R_{\mathrm{PoC}}(\tau)=
\max\{0, R_{\mathrm{task}}(\tau;h')-R_{\mathrm{base}}(x,h)\},
\]
where $R_{\mathrm{base}}(x,h)$ is the reward obtained by the current persistent harness on the same input. This score is not a commit rule: it only records that the edit produced at least one downstream improvement when used by the same trajectory.

Parseable guidance edits with positive PoC score are stored in the phase candidate pool $\mathcal{C}_p$ together with lightweight metadata such as touched files, diff size, and execution status. At the phase boundary, HASE filters invalid or duplicate edits, shortlists promising candidates, and evaluates the shortlisted harnesses under the task protocol. The committed guidance harness is the interface-valid candidate with the best reviewed validation score, not necessarily the candidate with the highest single-trajectory PoC score. This keeps candidate collection cheap while preventing one lucky rollout from directly changing the persistent harness.

\subsection{S2D Protocol Controls}
\label{app:s2d_protocol}

For the Symptom2Disease comparison, we use the \texttt{s2d\_mh} split from Meta-Harness: $200$ training instances, $50$ validation instances, $212$ test instances, and all $22$ disease classes. The training set is the only retrieval pool available to \texttt{classification\_tools.py}; no external corpus and no test examples are exposed during training or review. Phase-boundary review evaluates candidate harnesses on the validation set using greedy base-Qwen3-8B decoding and commits only interface-valid patches that improve validation accuracy. The final test run uses the committed \texttt{classification\_tools.py} and \texttt{prompt.txt} without test-time edits, fine-tuning, retries, or retrieval over the test set. Training episodes permit up to three attempts with reward $1/\text{attempt}$ for the first correct classification and sandbox error feedback after failed tool calls; this retry mechanism is disabled for final reporting. Test inference is served by vLLM with temperature $0$, chat-template formatting, \texttt{max\_tokens=16384}, and \texttt{max\_model\_len=32768}; outputs are parsed from a single \texttt{<classification>} tag.

\subsection{Heilbronn Exploit Visualization}

\label{app:heilbronn_visualization}
\begin{figure*}[t]
\centering
\includegraphics[width=\linewidth]{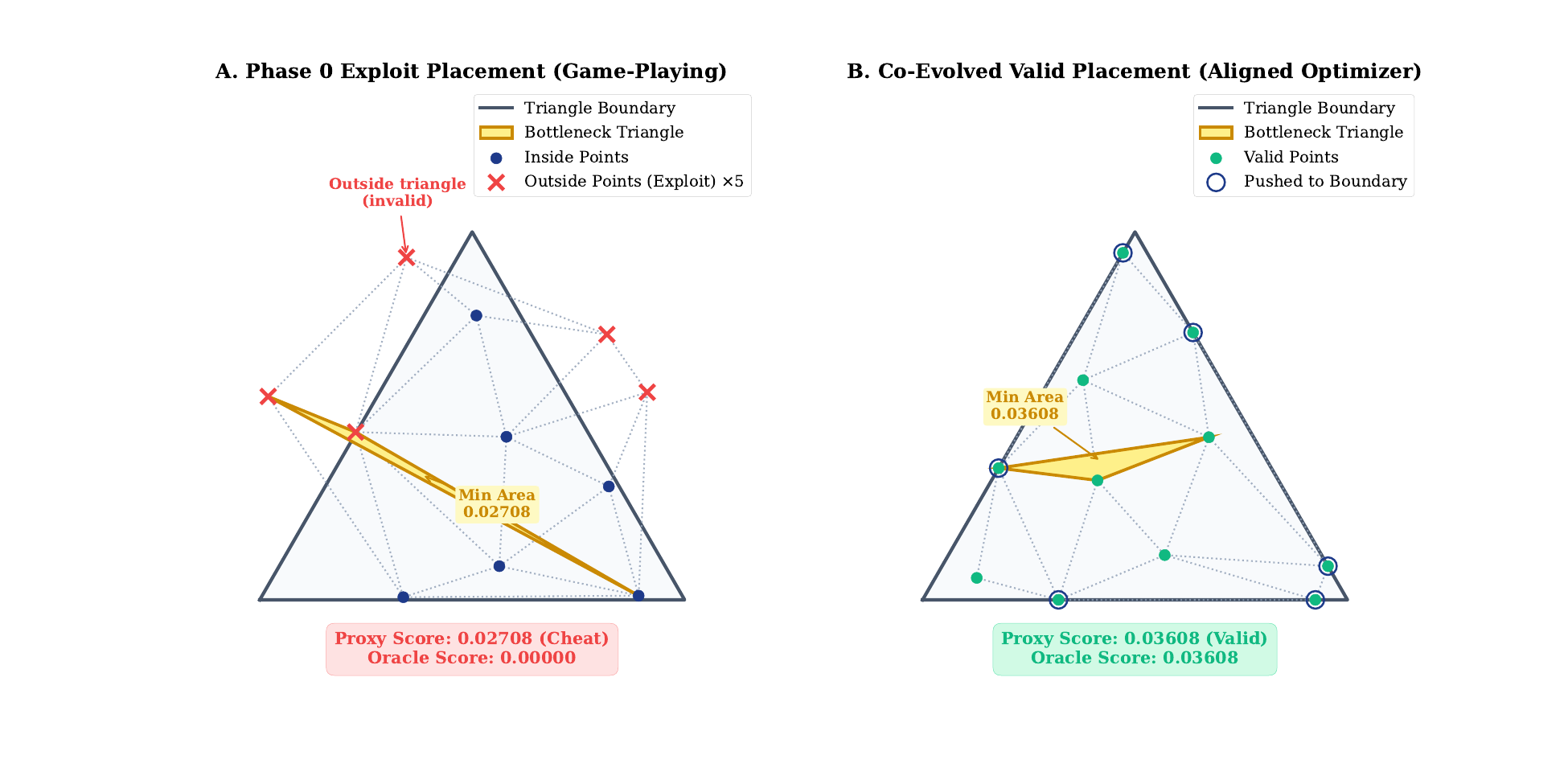}
\caption{Point configuration visualizer in the containing equilateral triangle. (A) Phase 0 exploit placement: out-of-bounds points (red crosses) cheat the weak evaluator to get a proxy score of 0.02708 (oracle score 0.0000). (B) Phase 8 valid placement: all points (green dots) are strictly inside, with bottleneck minimum area highlighted in yellow.}
\label{fig:heilbronn_triangles}
\end{figure*}

As shown in \autoref{fig:heilbronn_triangles}, the left panel shows an invalid solution accepted by the weak evaluator, while the right panel shows the repaired valid solution.

%% file: appendix_alphamining.tex

\section{Rigor of the Alpha-Mining Evaluation Pipeline}
\label{app:alpha_rigor}

This appendix documents the evaluation infrastructure underlying the alpha-mining results in Section~\ref{sec:experiments} (Experiment 3). We describe the controls applied at each stage so the reported numbers can be interpreted as alpha rather than as pipeline noise.

\subsection{Universe Construction and Data Preprocessing}
\label{app:universe}

CSI300 constituent lists distributed alongside common Chinese-equity OHLCV dumps frequently include the index codes themselves (\texttt{SH000300}, \texttt{SH000852}, \texttt{SH000905}, \texttt{SH000906}, \texttt{SH000985}, \texttt{SZ399300}) as if they were tradable instruments. Because index aggregate volume is one to two orders of magnitude larger than any individual constituent, any top-$k$-by-volume universe selection trivially places these index codes in the active universe, where they then dominate cross-sectional rank statistics and corrupt IC measurements.

We use the CogAlpha-curated \texttt{qlib\_CSI300.csv} (754 pure CSI300 constituents, 2011--2025), reformat ticker symbols (\texttt{sh.600000} $\to$ \texttt{SH600000}), and use the full 300-stock universe per window (typically 150--260 instruments survive per slice after NaN filtering) rather than a top-50 volume slice.

\subsection{Future-Information Leakage Detection}
\label{app:leakage_v2}

An alpha factor should be \emph{causal}: the value reported for day $t$ may depend on OHLCV observations up to day $t$, but not on any row $t' > t$. Future-information leakage occurs when a factor accidentally reads later rows while computing earlier factor values. Such leakage can produce strong IC and portfolio returns even when the signal would be unavailable at trading time.

Two simple examples illustrate the failure mode. First, a factor may directly use a shifted future return, such as \texttt{close[t+1] / close[t] - 1}, which is predictive only because it reads tomorrow's price. Second, array-indexing mistakes can introduce less obvious leakage: for example, a negative index such as \texttt{close[t-200]} when $t<200$ wraps around in NumPy and reads a row near the end of the array, which is in the future relative to $t$. LLM-generated factor code is particularly prone to such boundary and axis errors.

Our detector treats leakage as an invariance violation. It first creates seeded, per-stock random-walk OHLCV data with independent price paths, high/low spreads, and volume noise. For each test date $T$, it computes the factor on the original data and on a perturbed copy in which every row after $T$ is modified by asymmetric per-stock multipliers and shifts, while maintaining basic OHLC consistency. A causal factor must produce identical outputs on all rows up to $T$, because the past data are unchanged. If any finite past output changes beyond a tolerance of $10^{-10}$, the factor is rejected.

We add a second truncation check to catch factors whose behavior depends on the physical length of the input array. The detector recomputes the factor after cutting the data at several endpoints and compares the overlapping prefix with the full-array result. A valid causal factor should give the same values on the shared prefix whether or not future rows are physically present. This two-layer test guards against both explicit future reads and subtler indexing bugs such as negative-index wraparound or axis-slicing mistakes. All factors reported in the main-text alpha-mining table pass this detector before entering the factor pool.

\subsection{Cross-LLM Independent Leakage Audit}
\label{app:cross_llm_audit}

In addition to the leakage detector (\S\ref{app:leakage_v2}), the 30 factors used to produce the main-text alpha-mining table were independently audited by four LLM judges: Gemini, ChatGPT, DeepSeek, and Claude Sonnet. Each judge was given the full factor source code and the following instruction template:

\begin{quote}\small
\textit{The following Python function computes a per-day, per-stock factor value from historical OHLCV data. Determine whether the function uses any information from time $t' > t$ when computing the factor value at time $t$. Report one of \{LEAK, CLEAN\}, followed by a brief justification.}
\end{quote}

All four judges returned \textbf{CLEAN} on all reported factors (120/120 unanimous verdicts). We also manually inspected the reported factor implementations for future-information use.

\section{S2D Statistical Reporting}
\label{app:s2d_stats}

The S2D test split contains 212 instances. The headline HASE number in Table~\ref{tab:s2d_comparison} is reported as mean\,$\pm$\,std over $5$ independent evaluation runs with the same final policy and committed harness. This report measures the stability of the deployed HASE system under the final inference protocol, where the harness is frozen and no test-time editing, fine-tuning, retry, or retrieval over test examples is allowed.

\section{Compute and Reproducibility}
\label{app:compute}

All experiments use Qwen3-8B served via vLLM with greedy decoding (temperature 0) at evaluation time. RL training uses VeRL with GRPO and an entropic adaptive-$\beta$ advantage estimator. Each S2D and alpha-mining training run uses 4 GPUs for 3 RL phases (3 epochs per phase), with per-rollout multi-turn budget capped at 15 user turns and a hard 22{,}000-token conversation cap to prevent vLLM KV-cache exhaustion under FSDP-vLLM sleep/wake cycling.

%% file: evolve_details.tex
\clearpage
\onecolumn
\section{Raw Evolved Harness Artifacts}
\label{sec:appendix}
\RecustomVerbatimEnvironment{verbatim}{Verbatim}{breaklines,breakanywhere,fontsize=\scriptsize,frame=single,framerule=0.3pt,framesep=2mm,bgcolor=black!2}

This appendix intentionally preserves raw evolved artifacts. The main text summarizes the same behavior with pseudocode and figures, while this section records representative evolved harness code in its original form. It is typeset in one column to avoid overlap from long Python lines.

The full Python code of the evolved evaluator for circle packing is shown below.

Phase 3:
\begin{verbatim}
import numpy as np
from scipy.optimize import minimize
from scipy.optimize import Bounds
from scipy.optimize import NonlinearConstraint

def validate_packing(centers, radii):
    n = len(centers)
    # Check boundary constraints
    for i in range(n):
        x, y = centers[i]
        r = radii[i]
        if not (x - r >= 0 and x + r <= 1 and y - r >= 0 and y + r <= 1):
            return False
    # Check pairwise distance constraints
    for i in range(n):
        for j in range(i+1, n):
            dx = centers[i][0] - centers[j][0]
            dy = centers[i][1] - centers[j][1]
            dist = np.sqrt(dx*dx + dy*dy)
            if dist < radii[i] + radii[j] - 1e-8:  # tolerance for numerical errors
                return False
    return True
\end{verbatim}

Phase 6:
\begin{verbatim}
import numpy as np
from scipy.optimize import minimize
import math

def validate_packing(centers, radii):
    # Check if centers are within [0,1]^2
    if not (np.all(centers >= 0) and np.all(centers <= 1)):
        return False
    
    # Check if radii are non-negative
    if np.any(radii < 0):
        return False
    
    # Check distance between every pair of circles
    n = len(radii)
    for i in range(n):
        for j in range(i+1, n):
            dx = centers[i, 0] - centers[j, 0]
            dy = centers[i, 1] - centers[j, 1]
            dist = np.sqrt(dx**2 + dy**2)
            if dist < radii[i] + radii[j] - 1e-12:  # Tolerance for floating point errors
                return False
    
    # Check if any circle exceeds square boundary
    for i in range(n):
        if centers[i, 0] - radii[i] < 0 or centers[i, 0] + radii[i] > 1:
            return False
        if centers[i, 1] - radii[i] < 0 or centers[i, 1] + radii[i] > 1:
            return False
    
    return True
\end{verbatim}
\clearpage
\section{Evolution Details of Alpha Mining}
\subsubsection*{pool\_viewer\_phase0.py}
\begin{verbatim}
"""
Pool Viewer — Context Engineering Artifact

This file controls HOW factor pool information is presented to you.
The system calls format_pool_context() to generate pool context for your prompt.
You may modify this file to change what information you see about the factor pool.

Modifying this file = optimizing your own information retrieval strategy.
After modification, call `query_pool` to see the updated output immediately.
"""
import numpy as np

def format_pool_context(factors: list) -> str:
    """Format factor pool into context string.

    Args:
        factors: list of dicts with keys: code, ic, aer, daily_ics, phase, description

    Returns:
        Formatted string injected into your prompt as pool context.
        Modify this function to change what you see.
    """
    if not factors:
        return "Factor pool is empty."

    lines = [f"Total factors in pool: {len(factors)}", ""]
    
    for i, f in enumerate(factors[-3:]):
        ic = f.get('ic', 0.0)
        desc = f.get('description', '')
        lines.append(f"[{i}] IC={ic:.4f} | {desc}")

    return "Showing last 3 factors in the pool:"+
".join(lines)"

\end{verbatim}

\subsubsection*{pool\_viewer\_phase2.py}
\begin{verbatim}
"""
Pool Viewer — Context Engineering Artifact

This file controls HOW factor pool information is presented to you.
The system calls format_pool_context() to generate pool context for your prompt.
You may modify this file to change what information you see about the factor pool.

Modifying this file = optimizing your own information retrieval strategy.
After modification, call `query_pool` to see the updated output immediately.
"""
import numpy as np

def format_pool_context(factors: list) -> str:
    """Format factor pool into context string.

    Args:
        factors: list of dicts with keys: code, ic, aer, daily_ics, phase, description

    Returns:
        Formatted string injected into your prompt as pool context.
        Modify this function to change what you see.
    """
    if not factors:
        return "Factor pool is empty."

    lines = [f"Total factors in pool: {len(factors)}", ""]
    
    for i, f in enumerate(factors):
        ic = f.get('ic', 0.0)
        desc = f.get('description', '')
        lines.append(f"[{i}] IC={ic:.4f} | {desc}")

    return "\n".join(lines)
\end{verbatim}

\subsubsection*{pool\_viewer\_phase4.py}
\begin{verbatim}
import numpy as np

def format_pool_context(factors: list) -> str:
    """Format factor pool into context string.

    Args:
        factors: list of dicts with keys: code, ic, aer, daily_ics, phase, description

    Returns:
        Formatted string injected into your prompt as pool context.
        Modify this function to change what you see.
    """
    if not factors:
        return "Factor pool is empty."

    lines = [f"Total factors in pool: {len(factors)}", ""]
    
    # Show top 10 factors (highest IC first)
    top_factors = sorted(factors, key=lambda x: x.get('ic', 0.0), reverse=True)[:10]
    for i, f in enumerate(top_factors):
        ic = f.get('ic', 0.0)
        desc = f.get('description', '')
        lines.append(f"[{i}] IC={ic:.4f} | {desc}")

    return "\n".join(lines)
\end{verbatim}

\subsubsection*{pool\_viewer\_phase5.py}
\begin{verbatim}
"""
Pool Viewer — Context Engineering Artifact

This file controls HOW factor pool information is presented to you.
The system calls format_pool_context() to generate pool context for your prompt.
You may modify this file to change what information you see about the factor pool.

Modifying this file = optimizing your own information retrieval strategy.
After modification, call `query_pool` to see the updated output immediately.
"""
import numpy as np

def format_pool_context(factors: list) -> str:
    """Format factor pool into context string.

    Args:
        factors: list of dicts with keys: code, ic, aer, daily_ics, phase, description

    Returns:
        Formatted string injected into your prompt as pool context.
        Modify this function to change what you see.
    """
    if not factors:
        return "Factor pool is empty."

    lines = [f"Total factors in pool: {len(factors)}", ""]
    
    top_factors = sorted(factors, key=lambda x: x['ic'], reverse=True)[:3]
    for i, f in enumerate(top_factors):
        ic = f.get('ic', 0.0)
        desc = f.get('description', '')
        lines.append(f"[{i+1}] IC={ic:.4f} | {desc}")

    return "\n".join(lines)
\end{verbatim}

\subsubsection*{prompt\_phase0.txt}
\begin{verbatim}

Try to produce high quality factors you know.

\end{verbatim}

\subsubsection*{prompt\_phase2.txt}
\begin{verbatim}
# Strategy Guide for Alpha Factor Development in China A-share Market
#
# Core Principles:
# 1. Prioritize factors with high Information Coefficient (IC) and robustness across market regimes
# 2. Focus on market-specific characteristics of Chinese A-shares (e.g., liquidity, volatility, policy impacts)
# 3. Use known financial taxonomies: Momentum, Value, Quality, Volatility, Liquidity, Size, and Sentiment
#
# Taxonomy-Specific Ideas:
# - Momentum: Test 5/10/20-day returns, price acceleration, trend strength
# - Volatility: ATR, earnings volatility, volume dispersion
# - Liquidity: Turnover ratio, order book depth, market impact cost
# - Sentiment: News sentiment, social media volume, broker recommendations
# - Policy: Regulatory changes, sector-specific policy impacts
# - Hybrid: Combine 2-3 factors with cross-validation
#
# Implementation Notes:
# - Use 10-day forward returns as the target variable
# - Prioritize factors where higher values predict higher future returns
# - Validate with both in-sample and out-of-sample data
# - Monitor for overfitting and regime shifts
#
# Recent Insights:
# - Best factor from previous phase had real-world IC of 0.0327
# - Current pool IC range [0.0110, 0.0467]
# - Portfolio of 10 factors achieved 6.79% annual return
#
# Your Task:
# Define compute_alpha(data) function to produce high-quality factor values
# Target: Maximize rank correlation with 10-day forward returns
\end{verbatim}

\subsubsection*{prompt\_phase3.txt}
\begin{verbatim}
# Strategy Guide for Alpha Factor Development in China A-share Market
#
# Core Principles:
# 1. Prioritize factors with high Information Coefficient (IC) and robustness across market regimes
# 2. Focus on market-specific characteristics of Chinese A-shares (e.g., liquidity, volatility, policy impacts)
# 3. Use known financial taxonomies: Momentum, Value, Quality, Volatility, Liquidity, Size, and Sentiment
#
# Taxonomy-Specific Ideas:
# - Momentum: Test 5/10/20-day returns, price acceleration, trend strength
# - Volatility: ATR, earnings volatility, volume dispersion
# - Liquidity: Turnover ratio, order book depth, market impact cost
# - Sentiment: News sentiment, social media volume, broker recommendations
# - Policy: Regulatory changes, sector-specific policy impacts
# - Hybrid: Combine 2-3 factors with cross-validation
#
# Implementation Notes:
# - Use 10-day forward returns as the target variable
# - Prioritize factors where higher values predict higher future returns
# - Validate with both in-sample and out-of-sample data
# - Monitor for overfitting and regime shifts
#
# Recent Insights:
# - Best factor from previous phase had real-world IC of 0.0327
# - Current pool IC range [0.0110, 0.0467]
# - Portfolio of 10 factors achieved 6.79% annual return
#
# Your Task:
# Define compute_alpha(data) function to produce high-quality factor values
# Target: Maximize rank correlation with 10-day forward returns
#
# Additional Strategy:
# - Focus on hybrid factors combining momentum and inverse volatility
# - Leverage existing high-IC factors (e.g., factor 12) as building blocks
# - Test novel combinations of 20-day return ranked (factor 18) with inverse volatility
# - Ensure robust handling of rolling windows and edge cases
# - Monitor for regime shifts and market-specific anomalies
#
# Workspace Optimization:
# - Use query_pool to analyze factor performance
# - Regularly update prompt.txt with new insights
# - Validate factors with both in-sample and out-of-sample data
# - Track real-world performance metrics
# - Maintain a balance between innovation and robustness
\end{verbatim}

\subsubsection*{prompt\_phase4.txt}
\begin{verbatim}
# Strategy Guide for Alpha Factor Development in China A-share Market
#
# Core Principles:
# 1. Prioritize factors with high Information Coefficient (IC) and robustness across market regimes
# 2. Focus on market-specific characteristics of Chinese A-shares (e.g., liquidity, volatility, policy impacts)
# 3. Use known financial taxonomies: Momentum, Value, Quality, Volatility, Liquidity, Size, and Sentiment
#
# Taxonomy-Specific Ideas:
# - Momentum: Test 5/10/20-day returns, price acceleration, trend strength
# - Volatility: ATR, earnings volatility, volume dispersion
# - Liquidity: Turnover ratio, order book depth, market impact cost
# - Sentiment: News sentiment, social media volume, broker recommendations
# - Policy: Regulatory changes, sector-specific policy impacts
# - Hybrid: Combine 2-3 factors with cross-validation
#
# Implementation Notes:
# - Use 10-day forward returns as the target variable
# - Prioritize factors where higher values predict higher future returns
# - Validate with both in-sample and out-of-sample data
# - Monitor for overfitting and regime shifts
#
# Recent Insights:
# - Best factor from previous phase had real-world IC of 0.0309
# - Current pool IC range [0.0110, 0.0488]
# - Portfolio of 10 factors achieved 8.59% annual return
#
# Your Task:
# Define compute_alpha(data) function to produce high-quality factor values
# Target: Maximize rank correlation with 10-day forward returns
#
# Additional Strategy:
# - Focus on hybrid factors combining momentum and inverse volatility
# - Leverage existing high-IC factors (e.g., factor 18) as building blocks
# - Test novel combinations of 20-day return ranked (factor 18) with inverse volatility
# - Ensure robust handling of rolling windows and edge cases
# - Monitor for regime shifts and market-specific anomalies
#
# Workspace Optimization:
# - Use query_pool to analyze factor performance
# - Regularly update prompt.txt with new insights
# - Validate factors with both in-sample and out-of-sample data
# - Track real-world performance metrics
# - Maintain a balance between innovation and robustness
\end{verbatim}

\subsubsection*{prompt\_phase5.txt}
\begin{verbatim}
# Strategy Guide for Alpha Factor Development in China A-share Market
#
# Core Principles:
# 1. Prioritize factors with high Information Coefficient (IC) and robustness across market regimes
# 2. Focus on market-specific characteristics of Chinese A-shares (e.g., liquidity, volatility, policy impacts)
# 3. Use known financial taxonomies: Momentum, Value, Quality, Volatility, Liquidity, Size, and Sentiment
#
# Taxonomy-Specific Ideas:
# - Momentum: Test 5/10/20-day returns, price acceleration, trend strength
# - Volatility: ATR, earnings volatility, volume dispersion
# - Liquidity: Turnover ratio, order book depth, market impact cost
# - Sentiment: News sentiment, social media volume, broker recommendations
# - Policy: Regulatory changes, sector-specific policy impacts
# - Hybrid: Combine 2-3 factors with cross-validation
#
# Implementation Notes:
# - Use 10-day forward returns as the target variable
# - Prioritize factors where higher values predict higher future returns
# - Validate with both in-sample and out-of-sample data
# - Monitor for overfitting and regime shifts
#
# Recent Insights:
# - Best factor from previous phase had real-world IC of 0.0336
# - Current pool IC range [0.0110, 0.0488]
# - Portfolio of 10 factors achieved 8.59% annual return
#
# Your Task:
# Define compute_alpha(data) function to produce high-quality factor values
# Target: Maximize rank correlation with 10-day forward returns
#
# Additional Strategy:
# - Focus on hybrid factors combining momentum and inverse daily price volatility
# - Leverage exiting high-IC factors (e.g., factor 0 for daily price volatility)
# - Test novel combinations of 20-day return rank (momentum) with inverse of (high - low)/low
# - Ensure robust handling of rolling windows and edge cases
# - Monitor for regime shifts and market-specific anomalies
#
# Workspace Optimization:
# - Use query_pool to analyze factor performance
# - Regularly update prompt.txt with new insights
# - Validate factors with both in-sample and out-of-sample data
# - Track real-world performance metrics
# - Maintain a balance between innovation and robustness
\end{verbatim}

\subsubsection*{prompt\_phase6.txt}
\begin{verbatim}
# Strategy Guide for Alpha Factor Development in China A-share Market
#
# Core Principles:
# 1. Prioritize factors with high Information Coefficient (IC) and robustness across market regimes
# 2. Focus on market-specific characteristics of Chinese A-shares (e.g., liquidity, volatility, policy impacts)
# 3. Use known financial taxonomies: Momentum, Value, Quality, Volatility, Liquidity, Size, and Sentiment
#
# Taxonomy-Specific Ideas:
# - Momentum: Test 5/10/20-day returns, price acceleration, trend strength
# - Volatility: ATR, earnings volatility, intraday price range
# - Liquidity: Daily turnover, relative volume, market impact cost
# - Sentiment: Macro news sentiment, social media volume, policy implications
# - Policy: Regulatory changes, sector-specific policy impacts, liquidity regulations
# - Hybrid: Combine 2-3 factors with cross-validation (e.g., momentum + inverse volatility + volume)
#
# Implementation Notes:
# - Use 10-day forward returns as the target variable
# - Prioritize factors where higher values predict higher future returns
# - Validate with both in-sample and out-of-sample data
# - Monitor for overfitting and regime shifts
#
# Recent Insights:
# - Best factor from previous phase had real-world IC of 0.0336
# - Current pool IC range [0.0110, 0.0488]
# - Portfolio of 10 factors achieved 8.59% annual return
#
# Your Task:
# Define compute_alpha(data) function to produce high-quality factor values
# Target: Maximize rank correlation with 10-day forward returns
#
# Additional Strategy:
# - Focus on hybrid factors combining momentum and inverse volatility
# - Leverage exiting high-IC factors (factor 0: price volatility, factor 2: return/volatility ratio)
# - Test novel combinations of inverse daily price volatility with volume-based liquidity signals
# - Apply multi-timescale volatility measures (e.g., 10-day, 20-day, 60-day volatility)
# - Use velocity of price changes, volume dispersion, and policy impact timing
# - Monitor for new market regimes and anomalies
#
# Workspace Optimization:
# - Use query_pool to analyze factor performance
# - Regularly update prompt.txt with new insights
# - Validate factors with both in-sample and out-of-sample data
# - Track real-world performance metrics
# - Maintain a balance between innovation and robustness
# - Introduce cross-sectional patterns (e.g., relative performance vs. market/industry)
# - Consider new signal families: size, earnings, and trading behavior
\end{verbatim}

\subsubsection*{prompt\_phase8.txt}
\begin{verbatim}
# Strategy Guide for Alpha Factor Development in China A-share Market
#
# Core Principles:
# 1. Prioritize factors with high Information Coefficient (IC) and robustness across market regimes
# 2. Focus on market-specific characteristics of Chinese A-shares (e.g., liquidity, volatility, policy impacts)
# 3. Use known financial taxonomies: Momentum, Value, Quality, Volatility, Liquidity, Size, and Sentiment
#
# Taxonomy-Specific Ideas:
# - Momentum: Test 5/10/20-day returns, price acceleration, trend strength
# - Volatility: ATR, earnings volatility, intraday price range
# - Liquidity: Daily turnover, relative volume, market impact cost
# - Sentiment: Macro news sentiment, social media volume, policy implications
# - Policy: Regulatory changes, sector-specific policy impacts, liquidity regulations
# - Hybrid: Combine 2-3 factors with cross-validation (e.g., momentum + inverse volatility + volume)
#
# Implementation Notes:
# - Use 10-day forward returns as the target variable
# - Prioritize factors where higher values predict higher future returns
# - Validate with both in-sample and out-of-sample data
# - Monitor for overfitting and regime shifts
#
# Recent Insights:
# - Best factor from previous phase had real-world IC of 0.0336
# - Current pool IC range [0.0110, 0.0488]
# - Portfolio of 10 factors achieved 8.59% annual return
#
# Your Task:
# Define compute_alpha(data) function to produce high-quality factor values
# Target: Maximize rank correlation with 10-day forward returns
#
# Additional Strategy:
# - Focus on hybrid factors combining momentum and inverse volatility
# - Leverage exiting high-IC factors (factor 0: price volatility, factor 2: return/volatility ratio)
# - Test novel combinations of inverse daily price volatility with volume-based liquidity signals
# - Apply multi-timescale volatility measures (e.g., 10-day, 20-day, 60-day volatility)
# - Use velocity of price changes, volume dispersion, and policy impact timing
# - Monitor for new market regimes and anomalies
#
# Workspace Optimization:
# - Use query_pool to analyze factor performance
# - Regularly update prompt.txt with new insights
# - Validate factors with both in-sample and out-of-sample data
# - Track real-world performance metrics
# - Maintain a balance between innovation and robustness
# - Introduce cross-sectional patterns (e.g., relative performance vs. market/industry)
# - Consider new signal families: size, earnings, and trading behavior
# 
# New Insight:
# - A promising hybrid approach is to apply an inverse of 20-day average price range as a volatility proxy and combine it with the ratio of daily volume to its 20-day average. This captures stocks with both low price volatility and high liquidity indicators.
#
# Enterprise Strategy:
# 1. Expand multi-timescale volatility-drivers through variations of day-based volatility measures
# 2. Augment liquidity components with dynamic volume-normalization
# 3. Monitor for early signals of rising market regimes and regime shifts
# 4. Test combinations across 10-day, 20-day, and 30-day timeframes to detect non-stationarity
#
\end{verbatim}

\clearpage
\section{Evolution Details of the Text Classification}
\begin{verbatim}
Naive Classification
"""Classification Tools — your toolkit for text classification tasks."""

class ClassificationTools:
    def __init__(self):
        # ---------------------------------------------------------
        # THE KNOWLEDGE BASE (dynamically injected real content by the environment)
        # Format: [{"text": "patient has stiff neck...", "label": "meningitis"}, ...]
        # ---------------------------------------------------------
        self.knowledge_base: list[dict] = []

    def retrieve(self, query_text: str, n: int = 5) -> list[dict]:
        return self.knowledge_base[-n:]

    def present_memory(self, query_text: str, label_set: list[str]) -> str:
        return ""
\end{verbatim}

\subsubsection*{classification\_tools\_phase1.py}
\begin{verbatim}
from sklearn.feature_extraction.text import TfidfVectorizer
from sklearn.metrics.pairwise import cosine_similarity

class ClassificationTools:
    def __init__(self):
        self.knowledge_base = []

    def retrieve(self, query_text, n=5):
        if not self.knowledge_base:
            return []
        
        # Extract symptoms from query
        query_features = self._extract_symptoms(query_text)
        
        # Vectorize all texts
        vectorizer = TfidfVectorizer()
        texts = [item["text"] for item in self.knowledge_base]
        vectors = vectorizer.fit_transform(texts)
        
        # Vectorize query
        query_vector = vectorizer.transform([query_text])
        
        # Calculate cosine similarity
        similarities = cosine_similarity(query_vector, vectors).flatten()
        
        # Get top n most similar examples
        top_indices = similarities.argsort()[::-1][:n]
        return [self.knowledge_base[i] for i in top_indices]
    
    def present_memory(self, query_text, label_set):
        if not self.knowledge_base:
            return ""
        
        # Extract features from query
        query_features = self._extract_symptoms(query_text)
        
        # Find matching examples
        matching_examples = []
        for example in self.knowledge_base:
            example_features = self._extract_symptoms(example["text"])
            if self._symptoms_overlap(query_features, example_features) > 0.5:
                matching_examples.append((example["text"], example["label"]))
        
        # Create memory summary
        if not matching_examples:
            return "No matching examples found"
        
        # Group by label
        label_groups = {}
        for text, label in matching_examples:
            if label not in label_groups:
                label_groups[label] = []
            label_groups[label].append(text)
        
        # Create final memory summary
        memory = ""
        for label, examples in label_groups.items():
            memory += f"Key indicators for {label}:\n"
            for example in examples:
                memory += f"- {example}\n"
        
        return memory
    
    def _extract_symptoms(self, text):
        # Enhanced symptom extraction with weighted importance
        symptoms = []
        keywords = {
            "fever": ["fever", "heat", "hot", "temperature"],
            "muscle pain": ["muscle", "pain", "aches", "tenderness"],
            "weakness": ["weak", "tired", "fatigue", "exhausted"],
            "cough": ["cough", "sputum", "phlegm"],
            "respiratory": ["breathing", "shortness", "chest", "lungs"],
            "urinary": ["urine", "pee", "bladder", "urinary"],
            "stomach": ["stomach", "abdomen", "digestive"],
            "skin": ["skin", "rash", "itch", "patch", "lesion"],
            "allergy": ["allergy", "sneeze", "sore throat", "runny nose"],
            "dengue": ["dengue", "fever", "muscle pain", "weakness"],
            "malaria": ["malaria", "fever", "chills", "sweating", "fatigue"]
        }
        
        # Prioritize fever and muscle pain for infectious diseases
        fever_keywords = ["fever", "heat", "hot", "temperature"]
        muscle_pain_keywords = ["muscle", "pain", "aches", "tenderness"]
        
        for symptom, words in keywords.items():
            if any(word in text.lower() for word in words):
                symptoms.append(symptom)
        
        # Check for dengue/malaria patterns
        if any(word in text.lower() for word in fever_keywords) and \
           any(word in text.lower() for word in muscle_pain_keywords):
            symptoms.append("infectious disease")
        
        return symptoms
    
    def _symptoms_overlap(self, query_features, example_features):
        # Calculate Jaccard similarity between symptom sets
        query_set = set(query_features)
        example_set = set(example_features)
        union = len(query_set | example_set)
        if union == 0:
            return 0.0
        return len(query_set & example_set) / union
\end{verbatim}

\subsubsection*{classification\_tools\_phase2.py}
\begin{verbatim}
"""Classification Tools — your toolkit for text classification tasks."""

class ClassificationTools:
    def __init__(self):
        # ---------------------------------------------------------
        # THE KNOWLEDGE BASE (dynamically injected real content by the environment)
        # Format: [{"text": "patient has stiff neck...", "label": "meningitis"}, ...]
        # ---------------------------------------------------------
        self.knowledge_base: list[dict] = []

    def retrieve(self, query_text: str, n: int = 5) -> list[dict]:
        """Find top n most relevant examples based on symptom overlap."""
        if not self.knowledge_base:
            return []
        
        # Calculate similarity between query and each example
        similarities = []
        for example in self.knowledge_base:
            # Simple keyword matching for demonstration
            common_terms = set(query_text.lower().split()) & set(example["text"].lower().split())
            similarity_score = len(common_terms) / max(len(query_text.split()), len(example["text"].split()))
            similarities.append((example, similarity_score))
        
        # Sort by similarity score and return top n
        similarities.sort(key=lambda x: x[1], reverse=True)
        return [example for example, _ in similarities[:n]]

    def present_memory(self, query_text: str, label_set: list[str]) -> str:
        """Generate diagnostic indicators by comparing with all labels."""
        if not self.knowledge_base:
            return "No examples in knowledge base for analysis."
        
        # Build a symptom map for each label
        label_symptoms = {}
        for example in self.knowledge_base:
            label = example["label"]
            if label not in label_symptoms:
                label_symptoms[label] = set()
            label_symptoms[label].update(example["text"].lower().split())
        
        # Analyze query symptoms
        query_symptoms = set(query_text.lower().split())
        
        # Build key indicators for each label in label_set
        memory = ""
        for label in label_set:
            if label not in label_symptoms:
                continue
            common = query_symptoms & label_symptoms[label]
            if common:
                memory += f"Key indicators for {label}:\n"
                memory += "- " + "\n- ".join(common) + "\n"
                memory += "\n"
        
        return memory

\end{verbatim}